\documentclass[final,5p,times,twocolumn]{elsarticle}

\journal{Medical Engineering \& Physics}

\usepackage{epsfig}
\usepackage{amsmath}
\usepackage{hhline}
\usepackage{multirow}
\usepackage{float}
\usepackage{enumerate}
\usepackage{graphicx,url}
\usepackage{natbib}
\usepackage{graphicx}

\usepackage{setspace}
\begin{document}

\begin{frontmatter}

\title{Targeted Muscle Effort Distribution with Exercise Robots: Trajectory and Resistance Effects}

\author[First]{Humberto De las Casas} 
\author[First]{Santino Bianco}
\author[First]{Hanz Richter}

\address[First]{Mechanical Engineering Department, Cleveland State University, Cleveland, OH 44115 USA.}

\begin{abstract}
The objective of this work is to relate muscle effort distributions to the trajectory and resistance settings of a robotic exercise and rehabilitation machine. Muscular effort distribution, representing the participation of each muscle in the training activity, was measured with electromyography sensors (EMG) and defined as the individual activation divided by the total muscle group activation. A four degrees-of-freedom robot and its impedance control system are used to create advanced exercise protocols whereby the user is asked to follow a path against the machine's neutral path and resistance. In this work, the robot establishes a zero-effort circular path, and the subject is asked to follow an elliptical trajectory. The control system produces a user-defined stiffness between the deviations from the neutral path and the torque applied by the subject. The trajectory and resistance settings used in the experiments were the orientation of the ellipse and a stiffness parameter. Multiple combinations of these parameters were used to measure their effects on the muscle effort distribution. An artificial neural network (ANN) used part of the data for training the model. Then, the accuracy of the model was evaluated using the rest of the data. The results show how the precision of the model is lost over time. These outcomes show the complexity of the muscle dynamics for long-term estimations suggesting the existence of time-varying dynamics possibly associated with fatigue.
\end{abstract}
\end{frontmatter}
\noindent \textbf{Keywords:} Robotics; biomechanics; neural networks; muscle activation; control systems; sport physiology.\\
\noindent \textbf{Number of words in the document:} 3442.
\section{Introduction}
\begin{figure}[ht]
\begin{center}
\includegraphics[width=8cm]{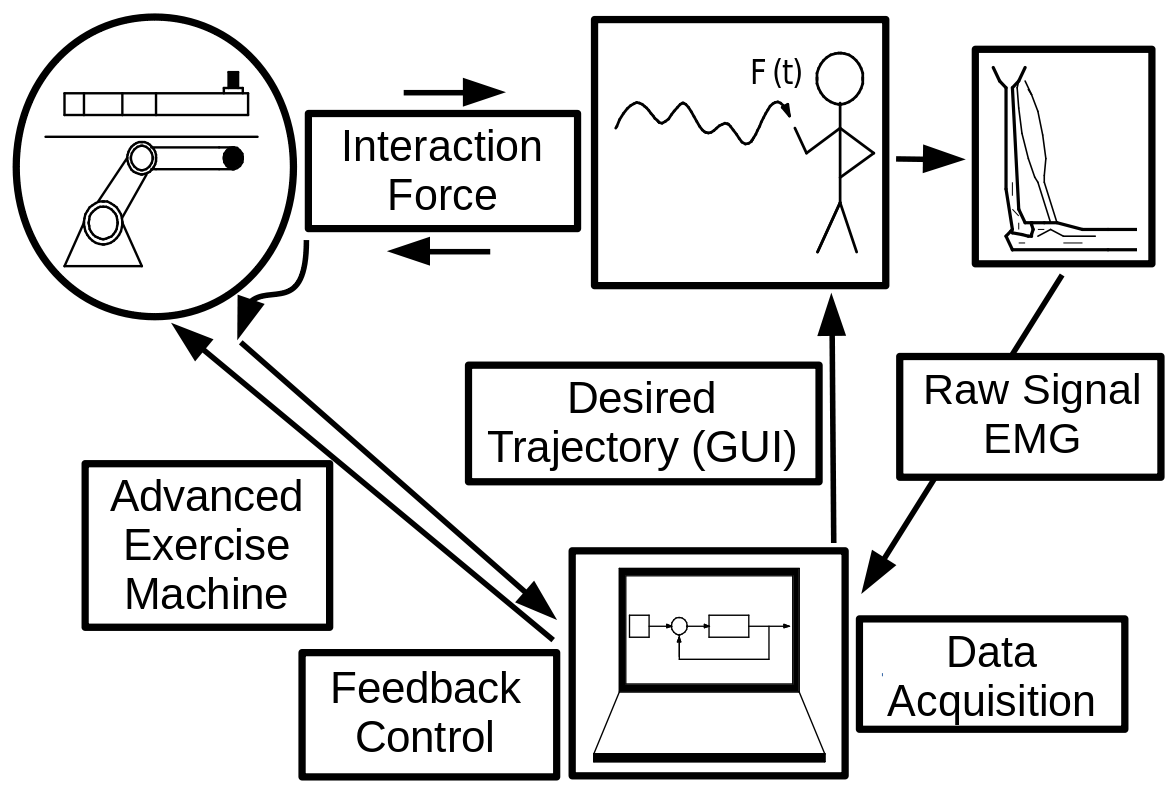}
\caption{HMI real-time experiment scheme.}
\end{center}
\label{Fig_Project_Scheme_0}
\end{figure}

\noindent The performance of human exercise depends on the trajectory and resistance settings of the exercise machine. Therefore, a relationship between these parameters and the muscle effort distribution as a result of the exercise is worth studying. In this work, we explore the relationship between muscle effort distributions and training parameters in advanced exercise protocols with a robotic exercise and rehabilitation machine. The muscular effort distributions, representing the participation of each muscle in the training activity, were estimated based on the muscle activations, and the training parameters used as variables during the exercise protocols were limited to the orientation of the ellipsoidal trajectory and the stiffness property of the impedance. This paper aims to determine this relationship by following an approach consisting of exercise experiments and model estimations. For the first part, experiments were carried out in human-machine interaction (HMI) environment (see Figure \ref{Fig_Project_Scheme_0}) to measure the muscle effort distribution as a result of the different combinations of robotic parameters. For the second part, the estimation was performed using a machine learning algorithm based on ANNs called feed-forward neural networks to identify time-varying dynamics altering the relationship over time.\\

\noindent Research in HMI using robots as advanced exercise machines (AEMs) has been widely explored in the last decade. Several applications in the areas of human performance and rehabilitation using these machines have been reported. Human performance research has made use of AEMs to enhance training conditioning by enabling innovative training patterns \citep{SPORTS_ROBOTS,MIPAPER1}. The area of rehabilitation has also seen recovery efficiency improvements using robots to help people with reduced motor skills \citep{MOT2,MOT3}. Moreover, several improvements in walking quality have been reported using a robotic prosthesis with smart controllers \citep{POYA2} leveraging advanced energy regeneration capabilities.\\

\noindent AEMs are similar to conventional training machines but use electric motors and control systems to provide controllable resistances and prescribed end-effector trajectories. These capabilities provide invaluable contributions to human performance and rehabilitation practices by intertwining exercise physiology with technology \citep{MOTOR}. Furthermore, the inherent capability of electric motors makes it possible to produce workloads even in microgravity \citep{ECC,ECC2}, showing potential benefits to overcome its detrimental effects such as muscle atrophy and bone mineral density (BMD) degradation \citep{ECC,NASA_2}. Furthermore, AEMs have great versatility for multi-target objectives for human performance and rehabilitation applications. The controller used for the AEM of this study is an impedance regulator. Impedance control is used to prescribe a desired dynamic relationship between the force applied to a mechanical system and its resulting motion \citep{hogan1985impedance,ROBUST2,ROBUST}. Impedance control can thus be used to produce damping or spring actions on systems that do not include physical damping or spring elements. Moreover, the perceived inertia of a mechanism can be reduced or increased, and the emulation of variable weights is possible.\\

\noindent The terms resistance and impedance are used interchangeably in this paper. Impedance controller is a good option for applications associated with advanced exercise and rehabilitation because of its capabilities to provide regulated resistances based on its parameters. The performance of these machines mainly lies in their trajectory and impedance parameters which can be easily manipulated. Previous research has reported the diversity of the physiological effects on the musculoskeletal, cardiovascular, and cardiorespiratory systems \citep{MIPAPER2} through the manual regulation of trajectory and impedance parameters using a powered training machine \citep{HUM_THE}. This research explores these training effects, but also the complex relationship between the effects and the machine parameters.\\

\noindent This work is about measuring and estimating parameters related to the physiology of human training. These parameters have been extensively studied, focusing mainly on muscle contractions, activations, and hypertrophy \citep{Measurements_MA_2,Measurements_MA_3}. For instance, some statistical methods have been developed to estimate action intervals by using electromyography (EMG) \citep{Measurements_MA_1,Estimation_HP_2} and some others for muscle forces and joint moments \citep{Estimation_HP_1}. These works have shown accurate estimations (less than 5\% error), however, for a rigorous study, the approach implemented for this study makes use of direct training measurements for data acquisition. Thus, the estimations conducted in this research have only been performed for the identification of time-varying muscle dynamics and by using artificial neural networks (ANNs). ANNs are used in many ways such as supporting tactical plannings during tournament games \citep{ANN_SPORTS_1}, predicting opponent performances based on nonlinear mathematical methods \citep{ANN_SPORTS_3}, and energy expenditure estimations during physical activity \citep{ANN_SPORTS_2}.\\

\noindent Robotics and information technology integration has been successfully used in an interactive environment to measure, support, and improve human exercise and rehabilitation practices \citep{EXP_SET_1,EXP_SET_2}. The objective of this technology is to enhance the manual and manipulative skills as they would otherwise do while playing a virtual reality game with their bodies \citep{EXP_SET_3}. One of the pioneers of robot-aided therapy systems is the MIT-Manus (MITM). This system has almost 30 years assisting stroke patients to improve their motor skills. MITM works by fitting the lower arm and wrist of the person into a brace attached to the arm of the robot. Meanwhile, he or she is required to follow a provided trajectory. The activity integration between human and robot assists the user by providing new neural connections which eventually support the muscle re-learning of the subject \citep{EXP_SET_4I,EXP_SET_4II}. From there, several extensions and new developments for MITM \citep{EXP_SET_5} have been performed, including using our 4OptimX robot. This robot provides a platform for research in the areas of human performance and rehabilitation.\\

\section{Experiment Setup}
\label{S_ExpSet}

\begin{figure}[ht]
\begin{center}
\includegraphics[width=6cm]{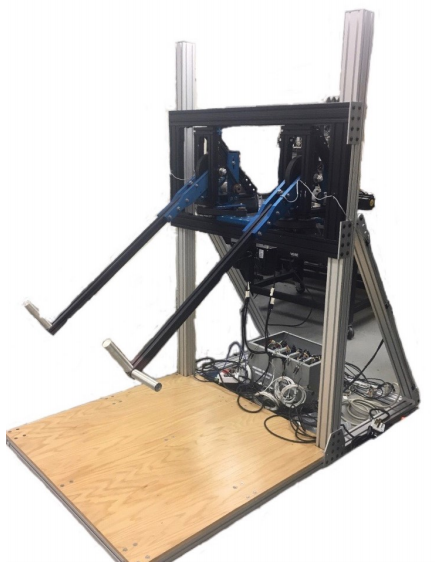}
\end{center}
\caption{4OptimX CSU robot.}
\label{Fig_4OptimX}
\end{figure}

\noindent The CSU 4OptimX robot (see Figure \ref{Fig_4OptimX}) is a 4 degrees-of-freedom (DOFs) robot consisting of two arms, each with 2 rotational DOFs with intersecting, perpendicular axes of revolution. Each end-effector (handle) thus moves on a  spherical surface. Each DOF is powered and controlled by an individual electric motor. The electric motors are torque-controlled. Together with a robust impedance control system, the motors effectively replace fixed weights, springs, and damping devices as means to produce resistance. Moreover, arbitrary impedances can be synthesized digitally by adjusting the control algorithm. Torque sensors are installed on each axis for use as feedback by the control system.\\

\noindent The controller for each DOF of the robot is a sliding mode impedance controller.  In the absence of any human-robot interaction (no external force), the controller tracks a reference trajectory. Any deviation from the reference position (as a result of manipulator interaction) imposes an impedance on each axis according to Equation \ref{equ_impedance_slidingmode}:  
\begin{equation}
\label{equ_impedance_slidingmode}
        \tau=I\ddot{e}+B\dot{e}+Ke
\end{equation}
\noindent where $e=x-x_d$ is the deviation between the actual position $x$ and the neutral (zero-effort) position $x_d$, 
$\tau$ is the torque produced on the motor due to the interaction force between the user and  the robot, and $I$, $B$ and $K$ are the inertial, damping and stiffness parameters, respectively.\\

\noindent The controller gives the system robust stability to unmodeled or inaccurately modeled plant dynamics and disturbances \citep{bianco2019robust}.  Stability is important to ensure the safety of the user during the articulation of the manipulator. Moreover, the controller's robustness ensures that the prescribed impedance is accurately achieved in the presence of inaccurately modeled or unmodeled plant dynamics.\\

\noindent Data acquisition, monitoring, recording, and real-time control of the robot is performed with the dSPACE MicroLabBox system (dSPACE GmbH, Paderborn, Germany), a Labjack T7 (LabJack Corporation, CO, USA), and a set of Trigno Wireless EMG sensors(Delsys Inc., MA, USA).  The Labjack was used to transmit information between the dSPACE and the computer displaying graphical data. The Delsys Trigno Wireless EMG system collected the EMG raw muscle activation signals and transmitted them to the dSPACE for recording.\\

\begin{figure}[ht]
\begin{center}
\includegraphics[width=3in]{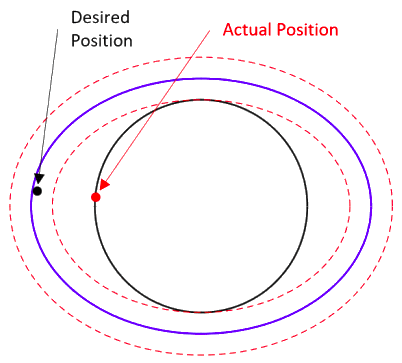}
\caption{GUI for the experiments.}
\label{fig_GUI_Impedance}
\end{center}
\end{figure}

\noindent The user is required to follow the desired position while receiving feedback from his or her current position. For this, a graphical user interface (GUI) is used to provide the required visual feedback (see Figure \ref{fig_GUI_Impedance}). The GUI consists of 4 curves and 2 dots. The blue curve is the ellipsoidal trajectory of fixed axis lengths and programmable orientation to be followed by the user. The black curve is the reference trajectory that is tracked by the robot in the absence of an external force. The 2 red dashed lines represent the tolerance limits where the actual position is suggested to remain during the performance of the experiment. The user's actual position (defined by the end-effector position of the robot) is labeled with the red dot. The desired position (rotating periodically over the blue ellipsoidal trajectory) is labeled with the black dot. The user position (red dot) is required to track the desired position (black dot) as best as possible but remain within the tolerance limits.\\

\noindent The trajectory and impedance settings of the exercise machine are expected to have a considered effect on the distribution of muscular effort \citep{TrainingPer1}. Stiffness and ellipse orientation were selected to evaluate their effects on the muscles since these parameters are expected to have a large influence. It is important to consider that in addition to these settings, organismic and intervening variables related to the musculoskeletal distribution, performance status, level of hydration, or mood seem to also produce an effect on the muscles. Furthermore, biological dynamics like muscle temperature or fatigue can potentially introduce time-varying dynamics. Therefore, muscle dynamics may be represented by the following function:
\begin{equation}
\label{M_fcn}
 \dot M=f(K,\theta,u(t)),
\end{equation}
\noindent where $M=[M_1,M_2,M_3,M_4,M_5,M_6]$ is a vector of muscle effort distributions, $K$ is the stiffness impedance, $\theta$ is the ellipse orientation, and $u(t)$ is the nonlinear time-varying function including all of the unknown and unmeasurable variables.\\

\noindent Throughout the experimental trials, the effort distribution, representing the muscle participation in the training activity, is estimated based on the muscle activations. The process begins with the EMG sensors producing raw signals which are real-time processed to generate muscle activations. Then, these activations are used to produce the muscle effort distribution. The complete conversion process from raw signal to muscle activation involves the following steps: the raw signals are recorded at a frequency of 2 kHz; a normalization is performed on the signals by removing the mean and dividing each value by its maximum activation (isometric test in trial 0); a second-order Butterworth band-pass filter between 30 and 950 Hz is used to clean the data; a full-wave rectification is implemented to convert the signal into only positive values; a second-order Butterworth low-pass filter at 50 Hz is used to attenuate the signal; and a normalization of each muscle activation is performed for the sum of all activations to obtain the muscle effort distribution as follows: 
\begin{equation}
    M=\frac{M}{\sum _{i=1}^{6}M_i},
\end{equation}
where $M$ is the same vector on Equation \ref{M_fcn}.

\section{Methodology}
\begin{figure}[ht]
\centerline{\psfig{figure=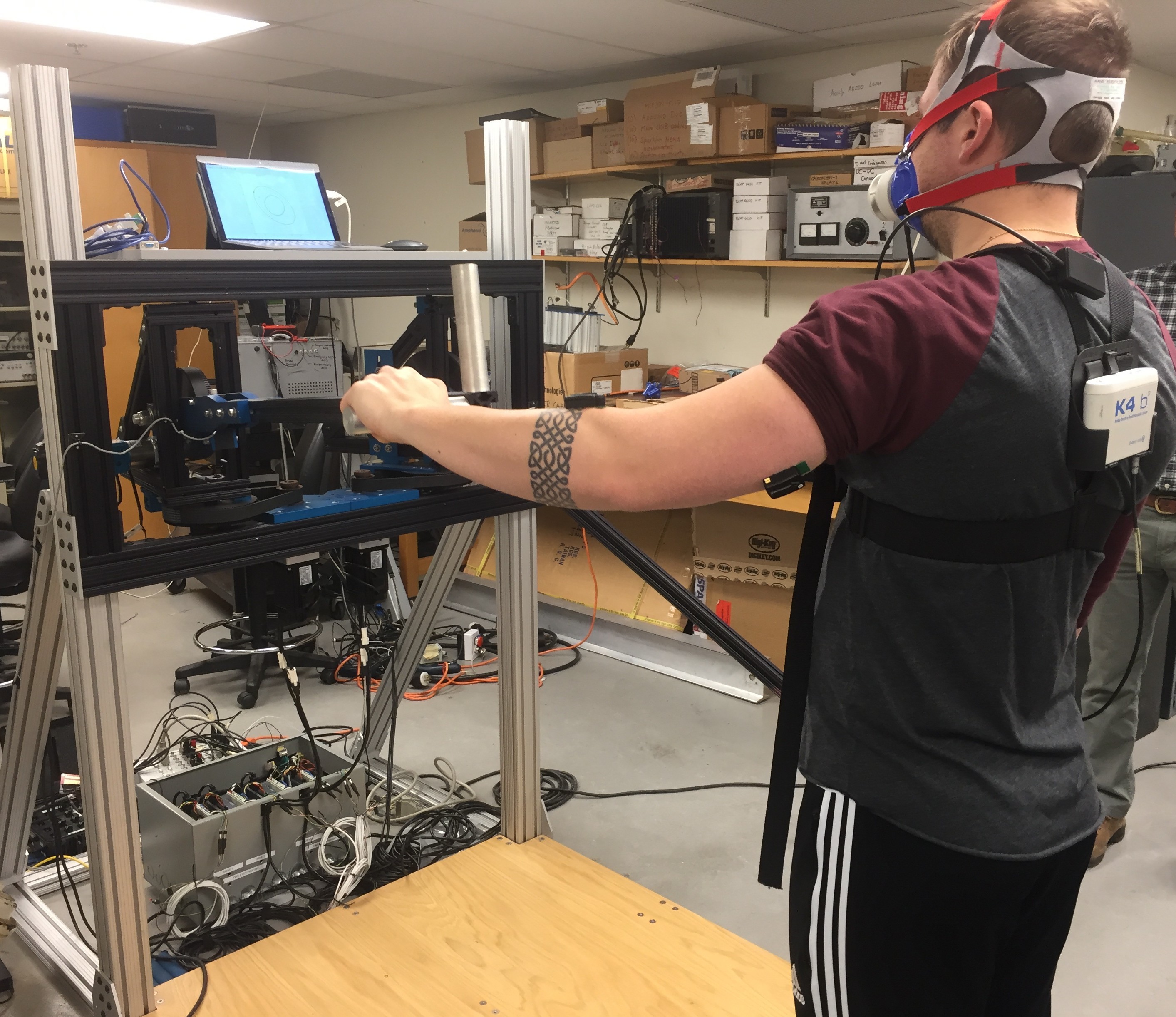,width=8cm}}
\caption{4OptimX experiment configuration.}
\label{fig_4OptimX_Exp}
\end{figure}

\noindent This research is divided into two parts. The first consisted of an experimental procedure to measure the muscle effort distribution as a result of multiple combinations of robotic parameters (see Figure \ref{fig_4OptimX_Exp}). The second consists of a model estimation of the robotic parameter combinations based on desired muscle distributions.\\

\subsection{Experimental Procedure}
\noindent The experimental procedure follows the conventional calibration process consisting, firstly, of a warm-up and isometric tests \citep{ISO2} (trial 0 in Table \ref{table_4OptimX1}). In addition to the warm-up, isometric tests are also used to assess muscle strength for the EMG sensor calibration. Thus, the subject is required to maintain a constant position where muscles are capable of producing maximum forces \citep{ISO1}. Then, the experiment proceeds with a 36-minute protocol including 17 individual trials with different combinations of robotic parameters (trials from 1 to 17 in Table \ref{table_4OptimX1}). Each trial was 1-minute long followed by a 1-minute rest period. The ``Low", ``High", and ``Super-high" labels for the impedances and speeds (in Table \ref{table_4OptimX1}) are defined in Tables \ref{table_4OptimX2} and \ref{table_4OptimX3}.\\

\begin{table}[ht]
\caption{4OptimX experiment planning (reference on Table \ref{table_4OptimX2} and \ref{table_4OptimX3}}
\begin{center}
\label{table_4OptimX1}
\begin{tabular}{c c c c c}
\hline
Trial & Impedance & Speed & Ellipse orientation \\ \hline
0 & \multicolumn{3}{c}{(Isometric test - EMG calibration)}\\
1 & Low &Low&90$^o$\\ 
2 & Low &Low&45$^o$\\
3 & Low &Low&0$^o$\\
4 & Low &Low&-45$^o$\\
5 & High &Low&90$^o$\\ 
6 & High &Low&45$^o$\\
7 & High &Low&0$^o$\\
8 & High &Low&-45$^o$\\
9 & High &High&90$^o$\\ 
10 & High &High&45$^o$\\
11 & High &High&0$^o$\\
12 & High &High&-45$^o$\\
13 & Low &High&90$^o$\\ 
14 & Low &High&45$^o$\\
15 & Low &High&0$^o$\\
16 & Low &High&-45$^o$\\
17 & Low &Super-high&0$^o$\\
 \hline
\end{tabular}
\end{center}
\end{table}

\begin{table}[ht]
\caption{Impedance reference for the 4OptimX experiment planning.}
\begin{center}
\label{table_4OptimX2}
\begin{tabular}{c c c}
\hline
Impedance & Low & High \\ \hline
Inertia (kgm$^2$/rad)& 0.035&0.035 \\
Damping (Nms/rad) &0.4&0.4 \\ 
Stiffness (Nm/rad) & 1&7 \\
\hline
\end{tabular}
\end{center}
\end{table}

\begin{table}[ht]
\caption{Speed reference for the 4OptimX experiment planning.}
\begin{center}
\label{table_4OptimX3}
\begin{tabular}{c c c c}
\hline
Speed & Low&High & Super-high \\ \hline
Period (s) &8 & 4 & 2 \\
\hline
\end{tabular}
\end{center}
\end{table}

\noindent A 22-year old male participant with a height of 180 cm and a weight of 91.8 kg was recruited. To participate in the study, the subject had to be free of any musculoskeletal injuries, cardiovascular disease, and/or any limitations that prevented him from participating in regular exercise. This subject was required to follow ellipsoidal trajectories with varying resistances acting at the end effector of the robot. Meanwhile, the subject's muscle activations were measured and recorded.\\

\begin{figure}[ht]
\centerline{\psfig{figure=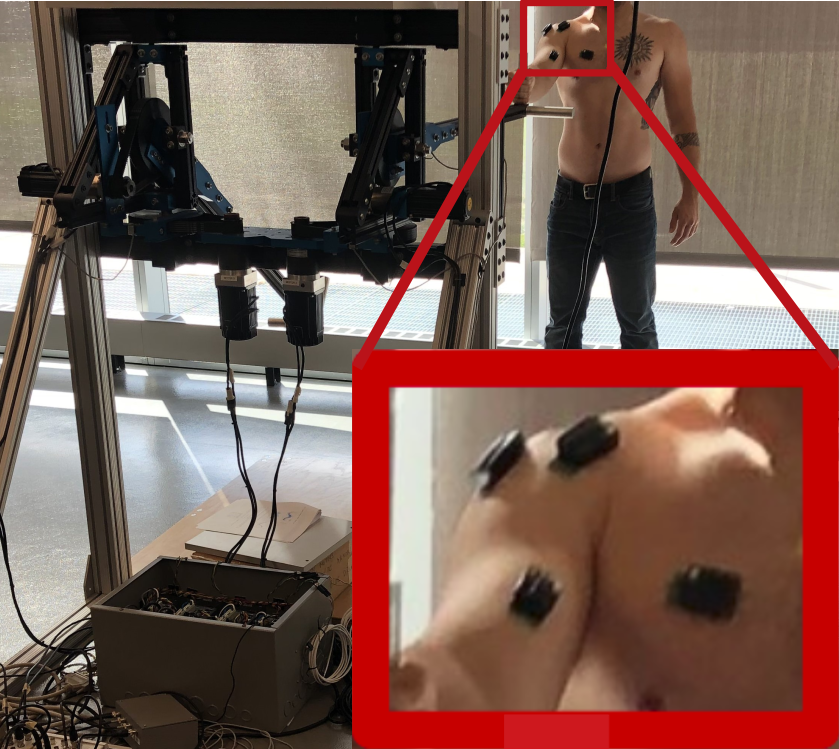,width=8cm}}
\caption{EMG location on the glenohumeral muscles.}
\label{fig_EMG_Muscles}
\end{figure}

\noindent The supplied trajectories were in various ellipsoid patterns (horizontal, vertical, and angled).  The major movements involved were flexion and extension in the sagittal plane, horizontal abduction and adduction in the transverse plane, and anterior circumduction movement. Therefore, some glenohumeral muscles \citep{Glenohumeral} (see Figure \ref{fig_EMG_Muscles}) were selected for the study in the following order: brachialis; posterior deltoid; anterior deltoid; biceps; triceps; and chest.\\

\noindent The brachialis and triceps, despite not belonging to the glenohumeral muscles, were chosen because of their relationship with the elbow movement. This relationship can provide information about involuntary rotations. The anterior and posterior deltoids were chosen because they are the main glenohumeral drivers. They are responsible for the space motion of the extended arm. The biceps brachii was chosen because of its synergistic work with the deltoid muscles. The chest was chosen because it is the main contributor to the glenohumeral adduction and stabilization of the shoulder. \\

\subsection{Robotic Parameters Estimation}
\noindent The objective of this part of the research is to perform a model estimation of the relationship between the robotic parameters and the muscle effort distribution. Since the last trial with the ``super-high" speed (trial 17) was only performed using one ellipse orientation, it will not be considered in the data set for the model estimation. Therefore, the remaining 16 previous trials were split into two parts. The first part of the data was used to train the model and the second part was used for testing. To observe the performance of the estimated model over time, in addition to the whole testing dataset (labeled as ``Whole"), the performance was also evaluated using the testing data split into 3 equally-sized sections labeled as ``First", ``Second", and ``Third".\\

\noindent The ANN algorithm was developed based on the feedforward neural network presented by Kubat \citep{MLBOOK}. Unlike in the experimental procedure, muscle effort distribution is was used as the input to estimate the required combination of the robotic parameters.\\   

\noindent Assuming that the muscle dynamics depend mainly on the external dynamics produced by the robot, the nonlinear time-varying term ($u(t)$) is neglected. Therefore, the muscle dynamics are re-defined as a function of only the robotic parameters as follows:\\
\begin{equation}
\label{M_ANN_fcn}
 \dot M=f(M,K,\theta),
\end{equation}

\begin{figure}[ht]
\centerline{\psfig{figure=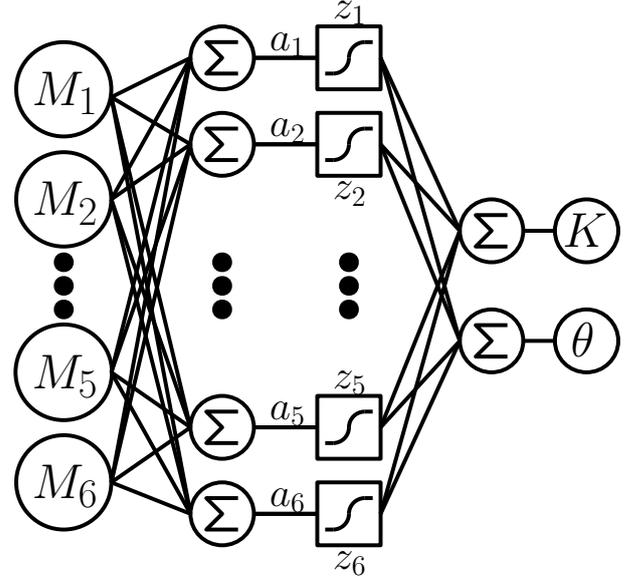,width=8cm}}
\caption{ANN Scheme.}
\label{fig_ANN}
\end{figure}

\noindent Based on Equation \ref{M_ANN_fcn}, the ANN scheme is built with 6 inputs (one per each muscle), 2 outputs (one per each robotic parameter), and a hidden layer with 6 nodes (selected experimentally). The complete scheme can be seen in Figure \ref{fig_ANN}. From this figure, the 2 internal ANN parameters, $a$ and $z$, are defined as follows:
\begin{equation}
a_i=\sum_{j=1}^6 M_jW_{in-ij},
\end{equation}
\begin{equation}
z_i=\frac{1}{1+e^{-a_i}},
\end{equation}
\noindent where $M$ is the vector of muscle effort distribution, $i$ is the hidden layer node, $j$ is the muscle number, and $W_{in-ij}$ is the value from the input weight
matrix of the order (i,j). The output is calculated as follows:
\begin{equation}
\hat{Y}=\sum_{i=1}^{6} z_i W_{out-ki},
\end{equation}
\noindent where $\hat{Y}$ is the estimated output vector $[\hat{K},\hat{\theta}]^T$, $k$ is the output number, and $W_{out-ki}$ is the value from the output weight matrix of the order (k, j).\\

\noindent The accuracy of the estimation relies on the calibration of the two weight matrices ($W_{in}$ and $W_{out}$) performed through data training. Thus, the calibration was performed using the training data and the backpropagation-of-error method recommended by Kubat \citep{MLBOOK}.\\

\section{Results}
\noindent The results from each part are presented independently. The first section presents the results from the 17 training trials. The second section presents the results from the robotic parameter estimations.

\subsection{Muscle Effort Distributions}
\noindent The activations of each muscle were averaged and plotted together to improve the visual comparison between trials. The individual muscle plots are presented together with muscle effort distribution. This presentation will show the variation of the muscle participation during each trial.\\

\begin{figure}[ht]
\begin{center}
\hspace*{-0.3cm}\includegraphics[width=9cm]{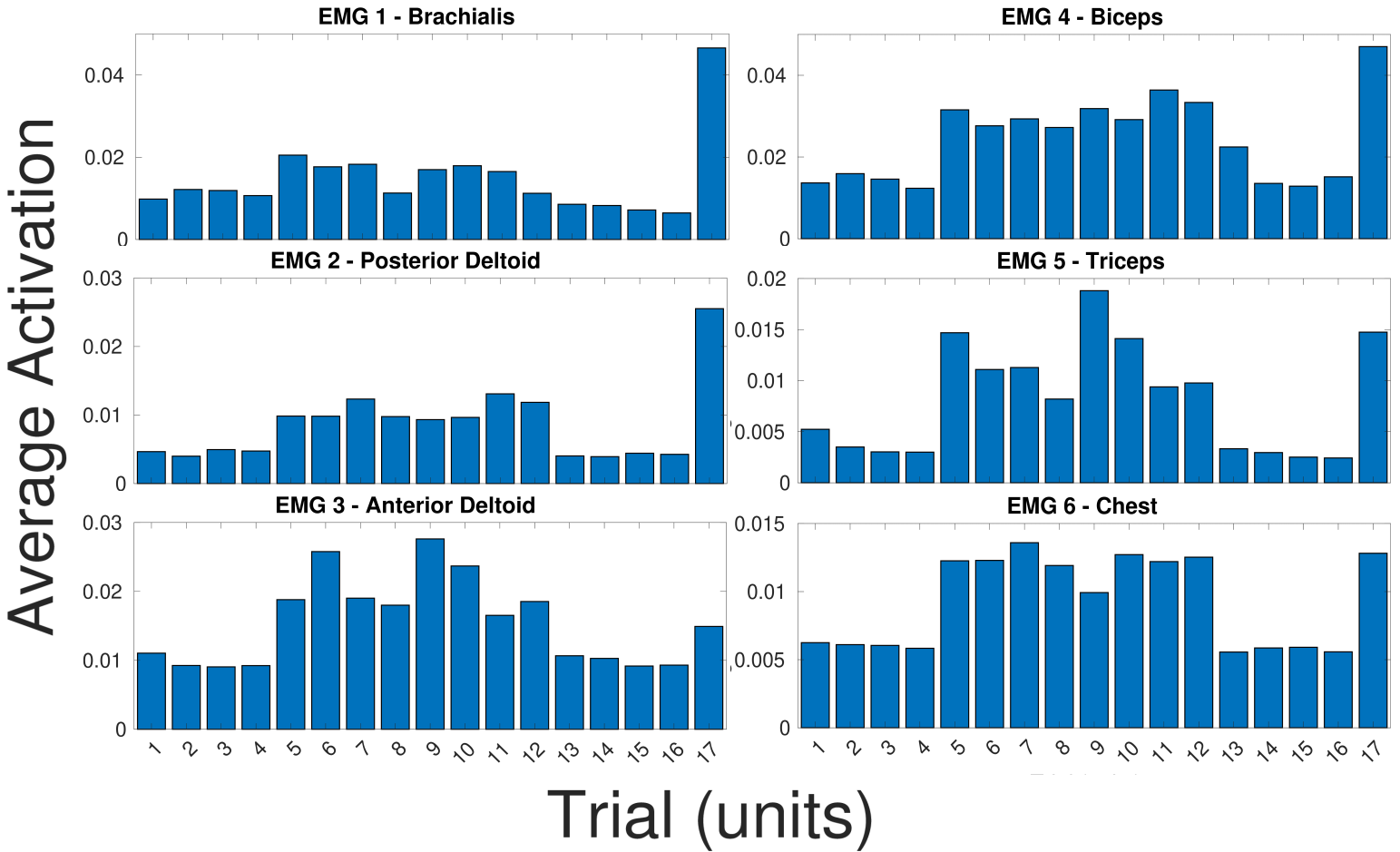} 
\caption{Average muscle activation on the brachialis (EMG-1), posterior deltoid (EMG-2), anterior deltoid (EMG-3), biceps (EMG-4), triceps (EMG-5), and chest (EMG-6) during each of the training trials.}
\label{fig_EMG_av}
\end{center}
\end{figure}

\begin{figure}[ht]
\begin{center}
\hspace*{-0.3cm}\includegraphics[width=9cm]{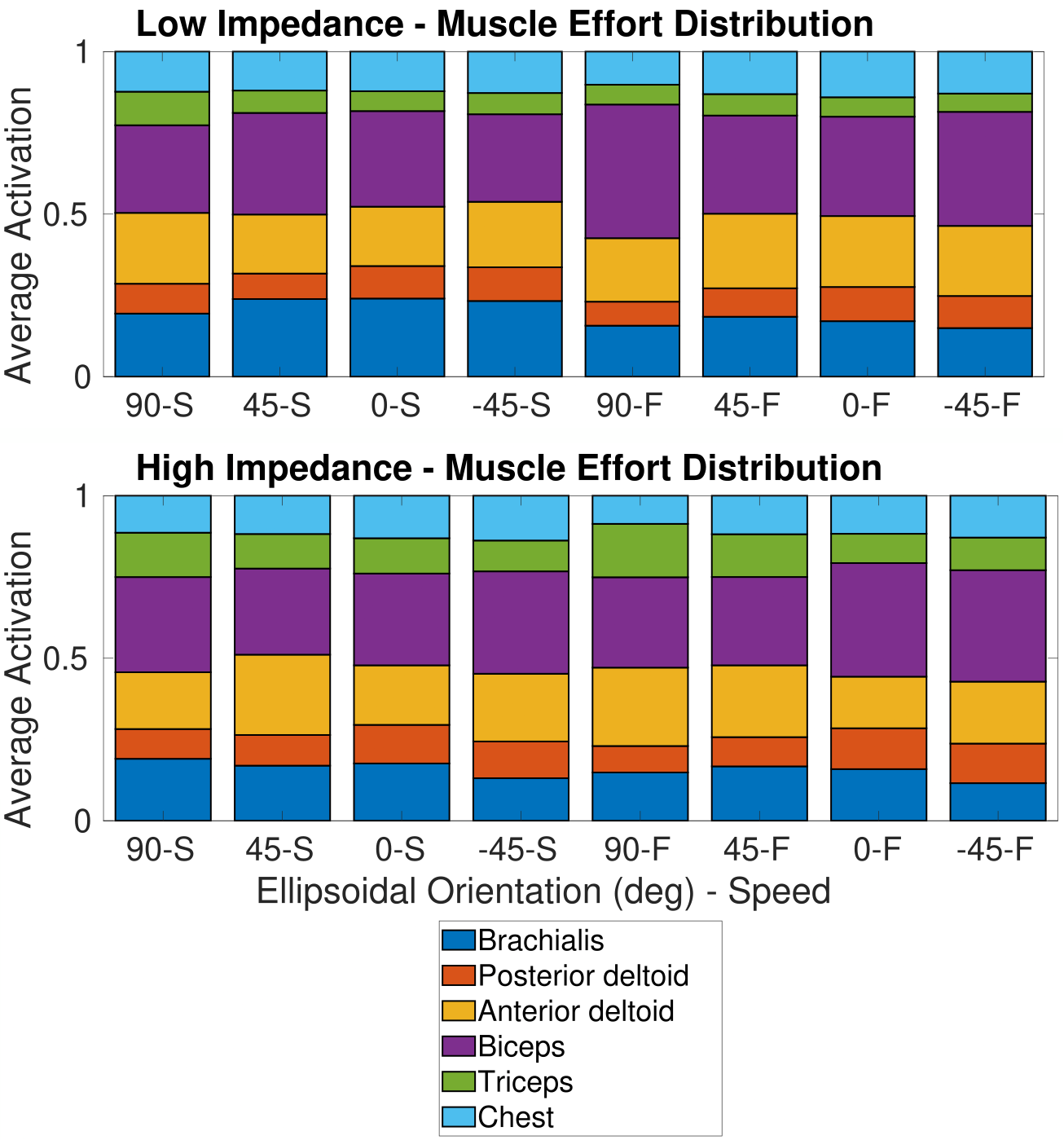} 
\caption{Average muscle effort distribution during the training trials with low (top plot) and high (bottom plot) impedances. The speed labels ``S" and ``F" refer to the trials with slow and fast frequencies respectively (periods of rotation of 8 and 4 seconds respectively).}
\label{fig_LH_MED}
\end{center}
\end{figure}

\noindent Analyzing the individual plots, high average activations on the brachialis, posterior deltoid, and biceps (see EMG numbers 1, 2, and 4 in Figure \ref{fig_EMG_av}) were seen during the last trial (17). This observation shows the significant effect of the very high circular trajectory frequency on the activation of these muscles. Nonetheless, it is important to consider that the frequency on the last trial was much higher than the other 16 trials (4 and 2 times faster than the slow and fast speed, respectively). A comparison between the slow trials (1-8) and the fast trials (9-16) does not show significant differences in the muscle activations.\\

\noindent On the other side, anterior deltoid, triceps, and chest (see EMG numbers 3, 5, and 6 in Figure \ref{fig_EMG_av}) showed a substantial effect on muscle activation due to the change in the impedance parameter. It seems that these muscles become the main drivers when the impedance is increased. Meanwhile, the change in the trajectory orientation was observed to be the most sensitive parameter. This could be attributed to the variation in the number of muscles used during each orientation. Every possible trajectory orientation can produce a different combination of muscle distributions.\\

\noindent Although the ellipse orientation was the most influential robotic parameter, the variation in the muscle effort distribution (see Figure \ref{fig_LH_MED}) saw rich and non-repeatable varieties of muscle effects with each parameter combination. In this way, the complexity of the effects of the robotic parameters on exercise performance is evidenced.\\

\subsection{Robotic Parameters Estimation}
\begin{figure}[ht]
\begin{center}
\hspace*{-0.3cm}\includegraphics[width=9cm]{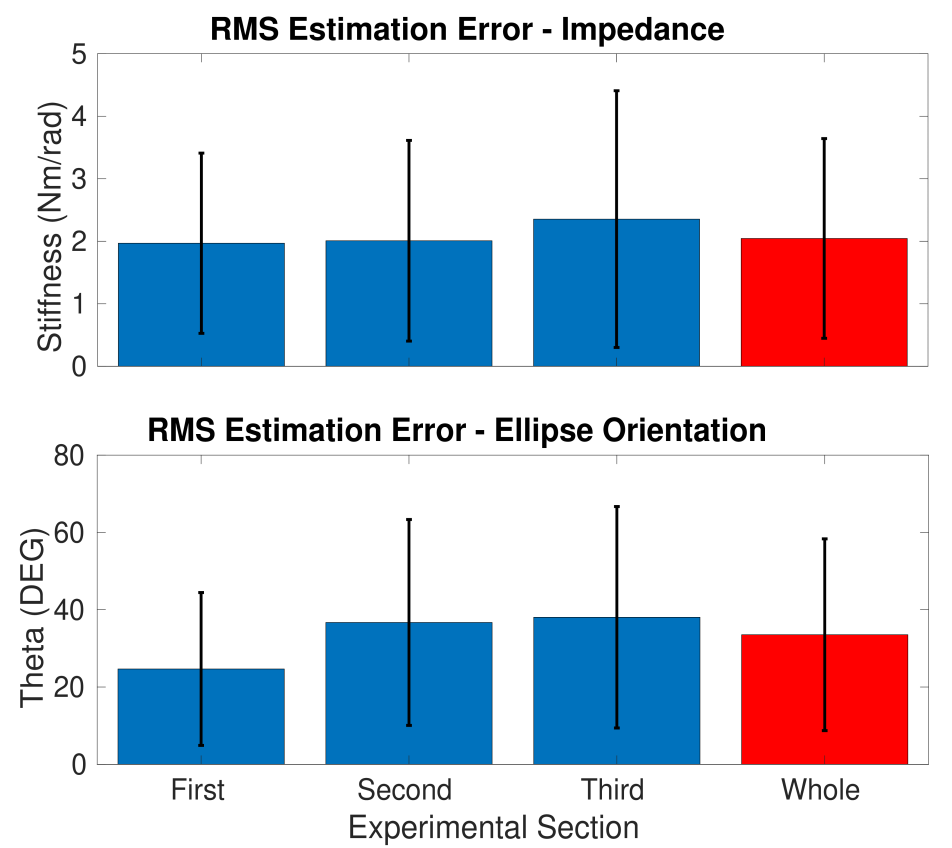} 
\caption{RMS error in the estimations of the impedance and the ellipse orientation parameters.}
\label{fig_RMS}
\end{center}
\end{figure}

\noindent After training the ANN using the backpropagation-of-error method, the testing data was used to measure the performance of the relationship estimation between the robotic parameters and the muscle effort distribution. The results from the root-mean-square (RMS) error between the real output and the estimated output are presented in Figure \ref{fig_RMS}).\\

\noindent Based on the similar values obtained from the split 3 sections and the whole testing data (see Figure \ref{fig_RMS}), the robotic parameters seems to have been efficiently estimated. However, it is important to highlight the fact that the best performance was achieved in the first section. For the impedance estimation, the RMS error went from 1.97 Nm/rad (section ``First") to 2.0 and 2.36 (for section ``Second" and ``Third", respectively). For the ellipse orientation, the RMS error went from 24.68 deg (section ``First") to 36.7 and 38.05 (for section ``Second" and ``Third" respectively). These results seem to show that the prediction accuracy of the ANN is lost with time, suggesting the presence of significant time-varying dynamics.\\

\noindent The time-varying dynamics affecting the muscles ($u(t)$ in Equation \ref{M_fcn}) were neglected due to modeling complexity. However, it seems to play an important role in the relationship with the robotic parameters. The time-varying dynamics are also dependent on biological factors associated with training such as fatigue, body temperature, and level of hydration. \\

\begin{figure}[ht]
\begin{center}
\includegraphics[width=8.5cm]{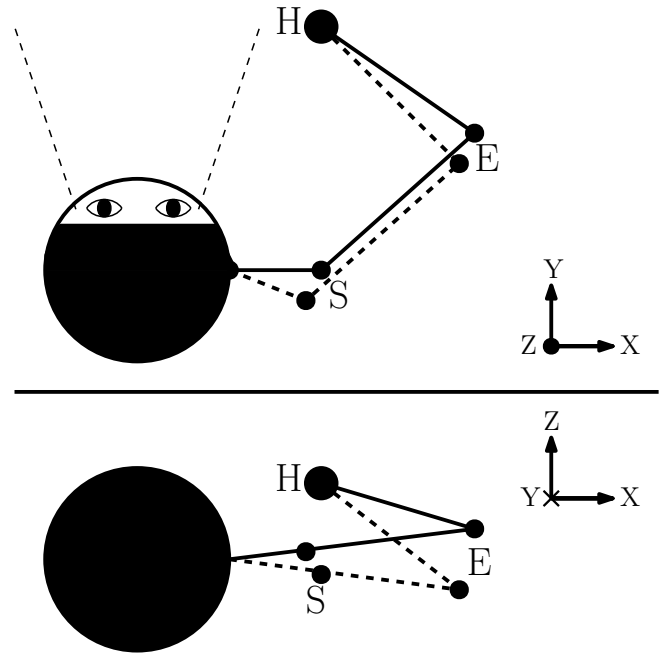} 
\caption{Example for the redundancy on the musculoskeletal orientation. (S, E, and H represents shoulder, elbow, and hand respectively).}
\label{fig_Multi_IK}
\end{center}
\end{figure}

\noindent The fully-actuated nature of the human body (having more muscle actuators than position variables) makes it possible to reach a target position with infinite possible musculoskeletal orientations (see Figure \ref{fig_Multi_IK}). Therefore, part of the changes in the model estimation could be attributed to the involuntary elbow rotations able to generate a completely new muscle effort distribution. These involuntary rotations could occur because of muscle fatigue or simply distraction.\\

\section{Conclusion and Future Work}\label{S_Future}
\noindent Upon completion of the experimental procedure, a great variety of muscle effort distributions displaying the diversity of the possible effects resulting from the robot parameters were witnessed. \\

\noindent Impedance and trajectory orientation not only showed a considerable effect over muscle activations but also reveal the potential for training versatility. Training parameters could be automatically set based on a muscle objective and several applications can be generated from that objective such as muscle group focus and/or isolation. The muscle objective can be professionally selected by a personal trainer or a therapist. For human performance applications, a personal trainer might emphasize or de-emphasize certain muscle groups forcing the user to train by using trajectories and impedances that maximize and/or minimize specific muscles. For rehabilitation applications, a therapist might target a specific muscle to focus on its particular recovery. Similarly, some parameters could be used to completely avoid the use of injured muscles without having to stop the training or rehabilitation procedures. Finally, levels of focalization could be introduced to increase the levels of training versatility. However, as previously noted, the involuntary inclusion of other muscles (while conducting experiments) potentially affects the model. Hence, an elbow immobilizer could be the solution to guarantee the achievement of a specific muscle objective.\\

\noindent It is important to consider that the trajectory tracking was not perfect but very consistent. The deviations between the desired position and the actual position were similar between each of the trials. A total RMS tracking error was reported at $0.2792\pm 0.0077$ (mean $\pm$SD) meters which were in the neighborhood of the individual trial values.\\ 

\noindent It was observed that under the same conditions of resistance and speed, higher activations were measured on the trials with an ellipsoidal orientation of 45$^o$ (trials: 5, 13, 21, and 29) and -45$^o$ (trials: 9, 17, 25, and 33). This suggests that these trajectories could potentially be considered for performance conditioning.\\


\noindent Future work will include optimization of the muscle effort distribution considering the time-varying dynamics observed in this study. It could be done by including the human dynamics inside of the closed-loop system. By this, enhancements in fitness and rehabilitation would be expected. For instance, training efficiency would increase, training environments would become safer, and recovery time would decrease.\\

\section*{Acknowledgements}
\noindent We wish to confirm that there are no known conflicts of interest associated with this publication and there has been no significant financial support for this work that could have influenced its outcome. This research was supported by the NSF, grant 1544702. This research has an IRB which covers Ethical Approval. It has been provided by Cleveland State University with the reference number 30305-RIC-HS.

\bibliographystyle{elsarticle-num}
\bibliography{elsarticle-template}

\end{document}